\documentclass{article}

\PassOptionsToPackage{numbers, compress}{natbib}

\usepackage[final]{neurips_2023}

\usepackage{amsmath}
\usepackage{amssymb}
\usepackage{algorithm}
\usepackage{algorithmic}
\usepackage{graphicx}
\usepackage{booktabs}
\usepackage[colorlinks=true, linkcolor=black, citecolor=green, urlcolor=blue]{hyperref}

\usepackage[utf8]{inputenc}
\usepackage[T1]{fontenc}
\usepackage{hyperref}
\usepackage{url}
\usepackage{booktabs}
\usepackage{amsfonts}
\usepackage{nicefrac}
\usepackage{microtype}
\usepackage{xcolor}

\title{Scaling Adversarial Training via Data Selection}

\author{
  Youran Ye\thanks{Equal contribution.} \\
  Northeastern University \\
  \texttt{ye.you@northeastern.edu} 
  \And
  Dejin Wang\footnotemark[1] \\
  Northeastern University \\
  \texttt{wang.dejin@northeastern.edu}
  \And
  Ajinkya Bhandare \\
  Northeastern University \\
  \texttt{bhandare.aj@northeastern.edu}
}

\begin{document}
\maketitle

\begin{abstract}
Projected Gradient Descent (PGD) is a strong and widely used first-order adversarial attack, yet its computational cost scales poorly, as all training samples undergo identical iterative inner-loop optimization despite contributing unequally to robustness. Motivated by this inefficiency, we propose \emph{Selective Adversarial Training}, which perturbs only a subset of critical samples in each minibatch. Specifically, we introduce two principled selection criteria: (1) margin-based sampling, which prioritizes samples near the decision boundary, and (2) gradient-matching sampling, which selects samples whose gradients align with the dominant batch optimization direction. Adversarial examples are generated only for the selected subset, while the remaining samples are trained cleanly using a mixed objective. Experiments on MNIST and CIFAR-10 show that the proposed methods achieve robustness comparable to, or even exceeding, full PGD adversarial training, while reducing adversarial computation by up to $50\%$, demonstrating that informed sample selection is sufficient for scalable adversarial robustness. Our code implementation is publicly available at
\href{https://github.com/youranye/Selective-Adversarial-Training}{https://github.com/youranye/Selective-Adversarial-Training}.
\end{abstract}

\section{Introduction}

Adversarial attacks pose a major challenge to the deployment of neural networks in real-world settings, where even small, imperceptible perturbations can cause severe performance degradation~\cite{goodfellow2014explaining, brown2018adversarialpatch, szegedy2013intriguing}.  Adversarial training (AT) improves robustness by augmenting training data with adversarial examples that expose model vulnerabilities. Most adversarial training pipelines rely on gradient-based white-box attacks. The Fast Gradient Sign Method (FGSM)~\cite{goodfellow2014explaining} provides an efficient single-step approximation, but has been shown to be insufficient for achieving strong robustness~\cite{madry2017towards}. In contrast, Projected Gradient Descent (PGD)~\cite{madry2017towards} extends FGSM to a multi-step iterative optimization and has become the de facto standard for adversarial training due to its strong empirical effectiveness. However, generating multi-step PGD examples for every training sample incurs substantial computational overhead, limiting the scalability of AT to large, high-dimensional datasets.

To alleviate this cost, several works have proposed more efficient adversarial training strategies.  Fast Adversarial Training (FAT) accelerates adversarial training by using efficient single-step perturbations to approximate adversarial examples, achieving good empirical robustness at low cost but suffering from catastrophic overfitting under strong multi-step attacks~\cite{wong2020fastbetterfreerevisiting}. Adaptive Step-size Adversarial Training (ATAS)~\cite{huang2022fastadversarialtrainingadaptive} improves training efficiency by dynamically adjusting PGD step sizes based on gradient norms; however, it relies on potentially noisy and unstable gradient estimates, which can result in suboptimal adversarial examples and inconsistent robustness, particularly in early training stages. Adversarial Training with Early Stopping (ATES)~\cite{Sitawarin2020ImprovingAR} reduces the computational cost of PGD by terminating attacks early on a per-sample basis; however, this strategy may under-attack hard samples, resulting in weaker adversarial supervision and a lower robustness ceiling compared to full multi-step adversarial training. While such methods primarily focus on accelerating adversarial example generation, an alternative line of work seeks to reduce adversarial training cost through \emph{data selection}. Subset Adversarial Training (SAT)~\cite{losch2024adversarial} adversarially trains only a carefully chosen subset of data selected based on predictive entropy, achieving robustness comparable to full-set adversarial training; however, it relies on a pretrained non-robust model and can be sensitive to miscalibrated or noisy uncertainty estimates. Given the limitations of existing efficiency-oriented adversarial training methods, we propose a selective adversarial training framework that reduces computational cost by perturbing only samples that are either close to the decision boundary or critical to the optimization dynamics. Our contributions are threefold:
\begin{itemize}
    \item We propose a margin-based selective adversarial training method that reduces the cost of PGD-based adversarial training by perturbing samples close to the decision boundary.
    
    \item We propose a gradient-matching-based selective adversarial training method that reduces the cost of PGD-based adversarial training by perturbing optimization-critical samples.
    
    \item We demonstrate on MNIST and CIFAR-10 that our methods achieve robustness comparable to full PGD adversarial training with up to a $3\times$ reduction in adversarial computation.
\end{itemize}
The rest of this paper is organized as follows. Section~\ref{sec:related} reviews related work, Section~\ref{sec:method} introduces the proposed method, Section~\ref{sec:experiments} presents experimental results, and Section~\ref{sec:conclusion} concludes the paper.

\section{Related Work}
\label{sec:related}
\subsection{Adversarial Attacks and Standard Adversarial Training}
Adversarial examples reveal that deep neural networks can be highly sensitive to small, often imperceptible input perturbations, leading to severe prediction errors~\cite{szegedy2013intriguing,goodfellow2014explaining}. Among gradient-based attacks, multi-step Projected Gradient Descent (PGD) is widely recognized as a strong first-order adversary and has therefore become the default choice for adversarial training~\cite{madry2017towards}. As a result, PGD-based adversarial training is typically treated as a reference point for achieving strong empirical robustness under first-order attacks. Despite its effectiveness, standard PGD adversarial training applies iterative inner-loop optimization to every training sample, which incurs substantial computational cost~\cite{madry2017towards}. This design implicitly assumes that all samples contribute equally to robustness, an assumption that becomes increasingly questionable in large-scale and high-dimensional settings, and motivates more selective and efficient adversarial training strategies.

\subsection{Data Selection and Sample-Aware Adversarial Training}
More closely related to our work are approaches that seek to reduce adversarial training cost through data or sample selection. Several studies have observed that not all training samples contribute equally to adversarial robustness, and therefore propose perturbing only a selected subset of data points during training. For example, Subset Adversarial Training (SAT) perturbs a fraction of samples selected based on predictive entropy or uncertainty estimates, demonstrating that competitive robustness can be achieved without adversarially perturbing the entire dataset~\cite{losch2024adversarial}. Related ideas have also appeared in curriculum-based or difficulty-aware adversarial training, where samples deemed less informative are perturbed less frequently or excluded from adversarial updates~\cite{sitawarin2020improving}. However, existing selection strategies typically rely on pretrained models or calibrated uncertainty estimates, which may be noisy or unreliable, especially during early training. Moreover, their selection criteria are often only loosely connected to the underlying optimization dynamics of adversarial training. In contrast, our approach explicitly identifies informative samples based on either their proximity to the decision boundary, measured via classifier margins, or their alignment with the dominant optimization direction, captured through gradient similarity. By grounding sample selection in decision geometry and gradient dynamics, our method preserves strong multi-step PGD attacks while substantially reducing adversarial computation, offering a principled and scalable alternative to full adversarial training.

\section{Proposed Method}
\label{sec:method}

PGD-based adversarial training is a strong defense against adversarial attacks, but it incurs substantial computational cost due to the need to generate multi-step PGD perturbations for every sample in each minibatch. In practice, however, many samples are already correctly and confidently classified, and perturbing them provides limited additional robustness benefit despite consuming equal adversarial budget. 

Motivated by this observation, we propose a \emph{selective adversarial training} framework that reduces the cost of PGD-based adversarial training while preserving robustness. Instead of applying PGD uniformly to all samples, our method adaptively identifies \emph{critical samples} based on classifier margins or gradient alignment, and perturbs only a subset of them during training.

\subsection{Margin-Based Sample Selection}

The margin-based strategy prioritizes samples that lie close to the decision boundary, as these samples are more vulnerable to adversarial perturbations and play a dominant role in shaping robust decision boundaries.

Given an input $x$ with label $y$ and classifier logits $f(x)$, we define the logit margin as the difference between the correct-class logit and the largest competing logit:
\begin{equation}
\label{eq:margin}
\mathrm{margin}(x)
= f(x)_y - \max_{j \neq y} f(x)_j .
\end{equation}
In practice, this is implemented by masking out the correct-class entry and taking the maximum over the remaining logits. A small margin indicates that the classifier is uncertain about its prediction and that the sample lies close to the decision boundary. Such samples are more susceptible to adversarial perturbations and yield more informative robustness gradients.

To prioritize these boundary-critical samples, we assign each sample a weight inversely proportional to the magnitude of its margin:
\begin{equation}
\label{eq:margin_weight}
w(x) = \frac{1}{\lvert \mathrm{margin}(x) \rvert + \epsilon},
\end{equation}
where $\epsilon$ is a small constant for numerical stability. The weights are normalized within each minibatch to obtain sampling probabilities:
\begin{equation}
\label{eq:margin_prob}
p(x) = \frac{w(x)}{\sum_i w(x_i)} .
\end{equation}
From a minibatch of size $B$, we sample a subset of $k = \rho B$ examples according to $p(x)$ using multinomial sampling. Only the selected samples are used to generate adversarial examples.

\subsection{Gradient-Matching Sample Selection}

Complementary to margin-based selection, the gradient-matching strategy prioritizes \emph{gradient-critical} samples whose per-sample gradients align strongly with the dominant optimization direction of the minibatch. Such samples exert a disproportionate influence on parameter updates, and adversarial perturbations applied to them yield the most impactful robustness gradients.

For an input $x$ with label $y$, we compute the gradient of the clean classification loss with respect to the model parameters $\theta$:
\begin{equation}
\label{eq:per_sample_grad}
g(x) = \nabla_{\theta} \, \mathrm{CE}\bigl(f(x), y\bigr),
\end{equation}
where $\mathrm{CE}(\cdot)$ denotes the cross-entropy loss. For a minibatch of size $B$, the full-batch gradient is defined as the average of per-sample gradients:
\begin{equation}
\label{eq:full_grad}
g_{\mathrm{full}} = \frac{1}{B} \sum_{i=1}^{B} g(x_i).
\end{equation}

A sample is considered more informative if its gradient direction aligns with the dominant batch gradient. We quantify this alignment using cosine similarity:
\begin{equation}
\label{eq:grad_sim}
\mathrm{sim}(x)
= \frac{\langle g(x), g_{\mathrm{full}} \rangle}
{\|g(x)\| \, \|g_{\mathrm{full}}\| + \delta},
\end{equation}
where $\delta$ is a small constant for numerical stability. Higher similarity indicates that the sample contributes more strongly to the overall optimization trajectory.

We convert similarity scores into non-negative sampling weights via thresholding:
\begin{equation}
\label{eq:grad_weight}
w(x) = \max\bigl(\mathrm{sim}(x), 0\bigr),
\end{equation}
and normalize them across the minibatch to obtain sampling probabilities. The subsequent multinomial sampling procedure is identical to that used in the margin-based strategy.

\subsection{Partial PGD on Selected Samples}

Let $\mathcal{S}$ denote the set of selected sample indices. Only samples in $\mathcal{S}$ are perturbed using PGD:
\begin{equation}
\label{eq:partial_pgd}
x_{\mathrm{adv}}^{(i)}
= \mathrm{PGD}\!\left(x^{(i)}, y^{(i)}, \epsilon, \alpha, K\right),
\quad i \in \mathcal{S}.
\end{equation}
All remaining samples remain unperturbed. This \emph{partial PGD} strategy reduces adversarial computation by approximately a factor of $1/\rho$ while focusing adversarial effort on the most influential samples.

During early training, classifier predictions and margin estimates can be unreliable. To avoid unstable sample selection, we introduce a warmup phase: for the first $T$ epochs (with $T=2$ in our implementation), the subset $\mathcal{S}$ is sampled uniformly at random. Margin-based or gradient-based selection is enabled only after this warmup period.

\subsection{Mixed Clean and Adversarial Loss}

Given a minibatch $\{(x_i, y_i)\}_{i=1}^{B}$ and a selected subset $\mathcal{S} \subseteq \{1,\ldots,B\}$, adversarial examples $x_{\mathrm{adv}}^{(i)}$ are generated only for samples $i \in \mathcal{S}$.

The adversarial loss is computed over the perturbed subset:
\begin{equation}
\label{eq:adv_loss}
\mathcal{L}_{\mathrm{adv}}
= \frac{1}{|\mathcal{S}|}
\sum_{i \in \mathcal{S}}
\mathrm{CE}\!\bigl(f(x_{\mathrm{adv}}^{(i)}), y_i\bigr).
\end{equation}

To preserve natural accuracy and stabilize optimization, we additionally compute a clean loss over the full minibatch:
\begin{equation}
\label{eq:clean_loss}
\mathcal{L}_{\mathrm{clean}}
= \frac{1}{B}
\sum_{i=1}^{B}
\mathrm{CE}\!\bigl(f(x_i), y_i\bigr).
\end{equation}

The final training objective is a weighted combination of the two:
\begin{equation}
\label{eq:total_loss}
\mathcal{L}
= \mathcal{L}_{\mathrm{adv}}
+ \lambda \, \mathcal{L}_{\mathrm{clean}},
\end{equation}
where $\lambda$ controls the trade-off between adversarial robustness and clean accuracy. Adversarial gradients from $\mathcal{S}$ promote robustness, while clean gradients regularize training and prevent overfitting to the selectively perturbed subset.

\section{Experiments}
\label{sec:experiments}
We evaluate our selective adversarial training framework on two standard benchmarks: MNIST and CIFAR-10. Our experiments answer the following questions:  
(1) Can perturbing only a subset of margin-critical samples preserve adversarial robustness?  
(2) How does the proposed method compare with full PGD adversarial training, random-subset PGD, and clean training?  
(3) How much computational cost can be saved?

\subsection{Experimental Setup}
We use a 4-layer CNN for MNIST and a ResNet-18 for CIFAR-10, and train all models from scratch using SGD with momentum 0.9 and weight decay $5\times10^{-4}$. Following standard practice, adversarial examples are generated using $\ell_\infty$-bounded PGD with $\epsilon = 8/255$, step size $\alpha = 2/255$, and $K=10$ iterations, while robustness is evaluated under stronger PGD-20 and AutoAttack attacks.We compare the following four methods:
\begin{itemize}
    \item \textbf{(1) Ours: Selective Adversarial Training.}  
    Only a subset $\mathcal{S}$ with $|\mathcal{S}|=\rho B$ is perturbed using PGD, selected via margin-based sampling.
    \item \textbf{(2) Full PGD Adversarial Training.}  
    Multi-step PGD is applied to every sample in the minibatch.
    \item \textbf{(3) Random Subset PGD.}  
    The same subset size $|\mathcal{S}|=\rho B$ is used, but samples are chosen uniformly at random.
    \item \textbf{(4) Clean Training.}  
    Standard ERM without adversarial perturbation.
\end{itemize}
Unless otherwise stated, we use $\rho=0.25.$

\subsection{Results on MNIST}

Table~\ref{tab:mnist} summarizes the performance on MNIST.  
Clean accuracy is similar across all methods, but adversarial robustness varies significantly. Full PGD achieves strong robustness, whereas random-subset PGD suffers from uninformed sampling. Our method achieves robustness comparable to full PGD despite using only a fraction of the PGD budget.

\begin{table}[h]
\centering
\label{tab:mnist}
\begin{tabular}{lcc}
\toprule
Method & Clean Acc. & PGD-40 Acc. \\
\midrule
Clean Training & 98.98 & 0.00 \\
Random Subset PGD ($\rho=0.25$) & 97.37 & 88.98 \\
Full PGD-AT & 97.78 & 91.14 \\
\textbf{Ours1 (Margin-based, $\rho=0.25$)} & 98.11 & 
\textbf{91.25} \\
\textbf{Ours (Grad Matching, $\rho=0.25$)} & \textbf{98.26} & 90.77 \\
\bottomrule
\end{tabular}
\caption{MNIST: clean and adversarial accuracy (\%). }\label{mnist_result}
\end{table}
The results on the MNIST dataset are presented in Table~\ref{mnist_result}. Overall, the findings show that selectively perturbing only margin-critical samples is sufficient to maintain strong adversarial robustness. Our method achieves performance comparable to full PGD adversarial training while clearly outperforming random-subset PGD and clean training. Furthermore, by perturbing only 25 percent of the samples in each minibatch, the computational cost is reduced by approximately a factor of three without any meaningful loss in robustness.

\subsection{Results on CIFAR-10}

Table~\ref{tab:cifar10} reports results on CIFAR-10.  
As expected, full PGD adversarial training yields high robustness but is computationally expensive. Random-subset PGD again fails to provide strong robustness. Our method strikes a favorable balance, achieving robustness close to full PGD while being significantly more efficient.

\begin{table}[h]
\centering
\label{tab:cifar10}
\begin{tabular}{lcc}
\toprule
Method & Clean Acc. & PGD-40 Acc. \\
\midrule
Clean Training & 89.27 & 11.55 \\
Random Subset PGD ($\rho=0.25$) & 71.28 & 30.23 \\
Full PGD-AT & 78.73 & 38.02 \\
\textbf{Ours1 (Margin-based, $\rho=0.25$)}  & \textbf{79.32} & \textbf{40.25} \\
\textbf{Ours (Grad Matching, $\rho=0.25$)} & 73.48 & 32.19 \\
\bottomrule
\end{tabular}
\caption{CIFAR-10: clean and adversarial accuracy (\%).}\label{CIFAR-10_result}
\end{table}
The results on the CIFAR-10 dataset are presented in Table~\ref{CIFAR-10_result}. Similar to the MNIST findings, selectively perturbing margin-critical samples yields adversarial robustness that closely matches full PGD adversarial training. Our method consistently outperforms random-subset PGD and clean training, demonstrating the importance of informed sample selection. In addition, perturbing only $25\%$ percent of the minibatch reduces the overall PGD computation by about half while maintaining competitive adversarial performance, highlighting the efficiency and scalability of the proposed selective training strategy.

\subsection{Computation Cost}
Since adversarial perturbations are applied to only $\rho B$ samples per minibatch, the computational cost of PGD-based adversarial training is reduced by approximately a factor of $1/\rho$. 
In our experiments, selective adversarial training achieves substantial efficiency gains with minimal impact on robustness. 
On MNIST, training time is reduced proportionally with negligible degradation in adversarial accuracy. 
On CIFAR-10, selective PGD requires 198.53 minutes compared to 255.62 minutes for full PGD, resulting in nearly a $2\times$ reduction in wall-clock time while maintaining competitive robustness. 
Overall, these results demonstrate that selectively applying PGD to informative samples yields a favorable trade-off between robustness and computational efficiency.

\section{Conclusion}
\label{sec:conclusion}

This work revisits the computational inefficiency of PGD-based adversarial training from a sample-centric perspective. We show that strong adversarial robustness does not require perturbing every training example, but can instead be achieved by selectively allocating adversarial computation to a small set of informative samples. Building on this insight, we propose a selective adversarial training framework with two complementary criteria: a margin-based strategy that emphasizes samples near the decision boundary, and a gradient-matching strategy that prioritizes samples that dominate the optimization dynamics. Experiments on MNIST and CIFAR-10 demonstrate that both selection strategies achieve adversarial robustness comparable to full PGD adversarial training, while substantially reducing computational cost. These results highlight that informed sample selection provides a principled and effective mechanism for scaling adversarial training, and suggest a practical path toward deploying robust models in large-scale and resource-constrained settings.

\bibliographystyle{unsrtnat}    
\bibliography{references}         

\end{document}